\documentclass[10pt, conference, compsocconf]{IEEEtran}

\IEEEoverridecommandlockouts

\usepackage{subcaption}
\usepackage{cite}
\usepackage[pdftex]{graphicx}
\usepackage{multirow}
\usepackage{array}
\usepackage{algorithmic}
\usepackage{algorithm}

\usepackage[cmex10]{amsmath}
\usepackage{amssymb}
\usepackage{url}
\usepackage{etoolbox}
\apptocmd{\thebibliography}{\scriptsize}{}{}

\newcommand{\upstar}{\textsuperscript{*}}
\newcommand{\upd}{\textsuperscript{$\dagger$}}
\newcommand{\updd}{\textsuperscript{$\ddagger$}}

\usepackage[super]{nth}

\begin{document}

\title{Language Model Supervision for Handwriting Recognition Model Adaptation}

\author{\IEEEauthorblockN{Chris Tensmeyer\upstar\thanks{\upstar{}Equal Contribution}\upd{}\updd{}, Curtis Wigington\upstar{}\upd{}\updd{}, Brian Davis\upd{}, Seth Stewart\upd{}, Tony Martinez\upd{} and William Barrett\upd{}}
\IEEEauthorblockA{
\upd{}Dept. of Computer Science -
Brigham Young University.
Provo, USA\\
\updd{}Adobe Research - San Jose, CA \\
tensmeyer@byu.edu}
}

\maketitle

\begin{abstract}
Training state-of-the-art offline handwriting recognition (HWR) models requires large labeled datasets, but unfortunately such datasets are not available in all languages and domains due to the high cost of manual labeling.
We address this problem by showing how high resource languages can be leveraged to help train models for low resource languages.
We propose a transfer learning methodology where we adapt HWR models trained on a source language to a target language that uses the same writing script. 
This methodology only requires labeled data in the source language, unlabeled data in the target language, and a language model of the target language. 
The language model is used in a bootstrapping fashion to refine predictions in the target language for use as ground truth in training the model.
Using this approach we demonstrate improved transferability among French, English, and Spanish languages using both historical and modern handwriting datasets.
In the best case, transferring with the proposed methodology results in character error rates nearly as good as full supervised training.
\end{abstract}


\section{Introduction}
State-of-the-art offline handwriting recognition (HWR) models are based on deep Convolutional Neural Networks (CNNs) and Bidirectional Long-Short Term Memory (BLSTM) networks and are trained on large amounts of labeled line images~\cite{graves2009novel}.
Obtaining such large annotated training sets is expensive and time consuming, because a person must segment thousands of text lines and manually type transcriptions for the ground truth.
However, such a process is often necessary for each language and domain because trained HWR models often fail to generalize sufficiently across domains, languages, and writers that were not observed during training.
Eliminating or lessening this requirement is the goal of unsupervised HWR and related approaches.

Prior work has attempted to address the lack of a large labeled training set for low resource languages/domains through several means.
Training on synthetic data is an appealing direction because an arbitrary amount of labeled data may be generated with little human effort.
In some works, synthetic data is obtained by applying annotation preserving transformations to real data in order to simulate the natural variability in handwriting~\cite{simard2003best,varga2008perturbation,wigington2017data}.
However, these methods depend on the availability of sufficiently diverse labeled data, which is not always the case.
Other works have modeled the writing process for generating isolated characters using prototypes for Chinese~\cite{tung1994performance} and Korean~\cite{lee1998new} characters, though it is not clear how such models could be extended to cursive scripts.
Elarian et al.~proposed a concatenative model for handwritten Arabic, though it relies on a database of pre-segmented characters and the concatenation procedure is specific to Arabic~\cite{elarian2015arabic}.

An alternative semi-supervised formulation of the problem assumes that there is a small labeled training set and a larger unlabeled training set.
The main methodology involves propagating annotations from the labeled set to the unlabeled set through model prediction.
Subsequent models then train as if the noisy predicted labels were ground truth annotations.
Frinken et al.~explored this method for isolated word image recognition in the framework of co-training, where a Hidden Markov Model (HMM) and a BLSTM model each made prediction that was used to further train the other model~\cite{frinken2011co}.
In a separate work, Frinken and Bunke use an ensemble of BLSTM networks for self-training, where high confidence ensemble predictions on the unlabeled data are subsequently used as ground truth to further train the ensemble.
Ball and Srihari used a similar idea to adapt writer specific HWR models from a general model by iteratively updating segmented character prototypes after performing recognition on unlabeled data~\cite{ball2009semi}.

In this work, we propose a transfer learning methodology that allows us to train a HWR model for a \emph{target} language for which we have no labeled images.
Our method only requires a labeled training set of line images in a sufficiently similar \emph{source} language, a trained Language Model (LM) in the target language, and a set of unlabeled images in the target language.
A source language is sufficiently similar to the target language if the character sets of the two languages have a large overlap. 
For example, Latin based languages, such as English, French and Spanish, are all sufficiently similar because they all use the written Latin script.
The LM can be obtained from digital text in the target language that is unrelated to the unlabeled images.
Digital text for training a LM is much more commonly available than labeled handwriting images, so our methodology helps extend automated HWR to lower resource languages.

After training a HWR model on the source language, our proposed method begins a hybrid training procedure where training occurs on both source data and target data.
There is no ground truth for the target data, so we combine the model prediction with the LM to produce a corrected prediction that we then use as ground truth.

We perform several experiments to analyze the behavior of our proposed transfer learning methodology for HWR.
These experiments are performed using 4 datasets and 3 languages: English, Spanish, and French.
We examine factors such as how long the source model is trained, LM decoding hyperparameters, and the proportion of source and target training used during hybrid training.
In the best cases, we find that transferring produces Character Error Rates (CER) nearly as low as those obtained by traditional supervised learning on the target data.

\section{Language Transfer Learning}

We formulate our problem as follows.
Suppose we wish to obtain a trained HWR model for a target language Y that has no labeled training data available, but there are many unlabeled text line images in this language.
We also have sufficient digital text in language Y, such that we can train a Language Model (LM).
For another language X we have segmented text line images with corresponding ground truth transcriptions.
Noting that languages X and Y have similar character sets, we want to use the data in both languages to produce the HWR model for language Y.

Though our discussion uses the term \emph{language}, our methodology is also applicable to transfer learning HWR problems where there is a difference in domains (e.g. modern vs historical) or writers.
We demonstrate this later by transferring between a modern English dataset and a historical English dataset.

\subsection{Source Model Training}

\begin{figure}
\centering
\includegraphics[width=0.35\textwidth]{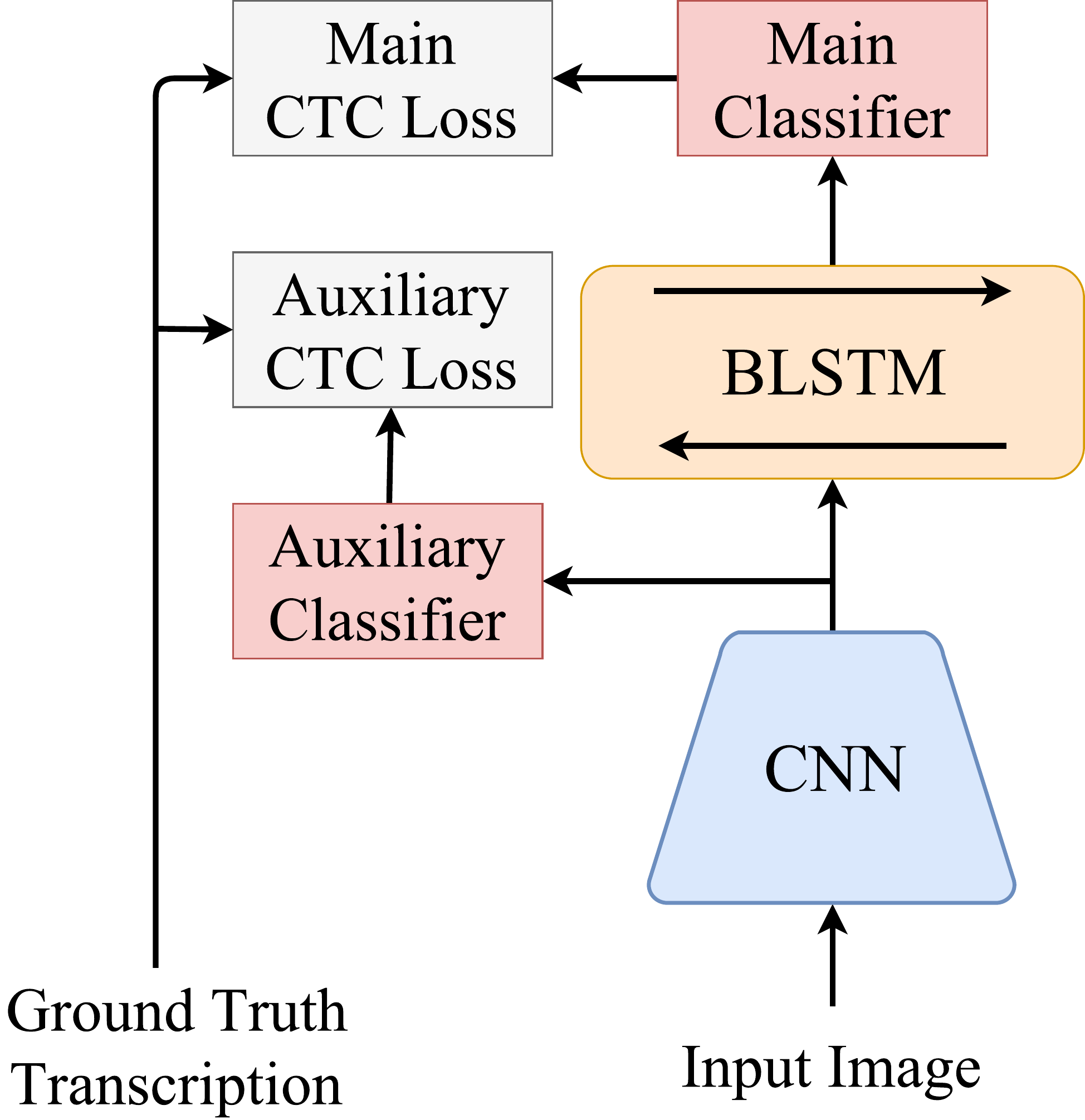}
\caption{CNN-BLSTM architecture with auxiliary classifier and loss after the CNN.}
\label{fig:arch}
\end{figure}

We begin transfer learning by training a state-of-the-art HWR model on the source language for which we have ground truth transcriptions.
Fig.~\ref{fig:arch} shows our CNN-BLSTM architecture, which is similar to the model in~\cite{wigington2017data}, but introduces an auxiliary classifier and loss.
This model learns high level features using convolution operations that are vertically collapsed to a 1D horizontal sequence of feature vectors that are fed to a 2-layer BLSTM.
In the BLSTM, context is propagated both forwards and backwards along the sequence.
Two separate frame-wise, linear character classifiers are each applied to the output of the CNN and the output of the combined CNN-BLSTM.
Both classifiers are trained using Connectionist Temporal Classification (CTC) loss~\cite{graves2006connectionist} which automatically aligns frame-wise outputs with the ground truth transcriptions.

The classifier that operates on the output of the CNN is considered an auxiliary classifier and it is discarded after the training procedure, meaning that the model outputs the predictions made by the main classifier that operates on the output of the BLSTM.
We found that introducing an auxiliary classifier improves transferability of the model, likely because it forces the CNN visual features to be discriminative of characters themselves instead of depending on further processing from the BLSTM layers.
When transferring between languages, the visual difference of some shared characters is small, so the CNN should be robust to the language difference.
In contrast, the BLSTM considers the whole sequence, so it is more sensitive to transferring between datasets.

The precise architecture of our HWR model is based on the model presented in~\cite{wigington2017data}.
The size of the the input image is $W \times 60$, where $W$ is the image width, which can dynamically vary.
The CNN is composed of 6 convolutional layers with 3x3 learnable kernels, and there are 64, 128, 256, 256, 512, and 512 feature maps respectively for the 6 layers.
We apply Batch Norm (BN) after layers 4 and 5 and 2x2 Max-Pooling (MP) with a stride of 2 after layers 1 and 2.
After layers 4 and 6, we vertically collapse features by using 2x2 MP with a vertical stride of 2 and a horizontal stride of 1.
To form the input for the BLSTM and for the CNN auxiliary classifier, we concatenate features in the same column to form a 1D horizontal sequence of 1024-dimensional feature vectors.
The BLSTM has 2-layers each with 512 hidden nodes that have a 0.5 probability of node dropout.
A linear classifier is applied to each time step to produce the final prediction, which is a probability distribution over characters at each timestep.

The model is trained using CTC loss over both the main classifier and the auxiliary classifier:
\begin{equation} \label{eq:two_loss}
L(x,y) = \lambda L_{CTC}(\phi(x),y) + (1 - \lambda) L_{CTC}(\psi(x), y)
\end{equation}
where $x$ represents an input image, $y$, the corresponding ground truth transcription, $L_{CTC}$ is the CTC loss~\cite{graves2006connectionist}, $\phi$ is the auxiliary CNN classifier, and $\psi$ is the BLSTM classifier. 
We empirically set $\lambda=0.25$ based on cross validation using validation data.

\subsection{Language Model Decoding}

The HWR model predicts each output character independently, and this may produce linguistically improbable sequences of characters.
Decoding with a Language Model (LM) combines the individual predicted character probabilities with how likely sequences of characters occur together.

Our LM decoding implementation uses the Kaldi Speech Recognition Toolkit~\cite{Povey_ASRU2011}, which has been used previously in HWR models~\cite{bluche2014comparison}.
Similar to~\cite{voigtlaender2016handwriting}, we use a 10-gram character LM, which estimates $p(c_i|c_{i-1}c_{i-2}\dots c_{i-9})$ from digital text.
Not all 10-gram character sequences are observed, so we smooth the empirical 10-gram distribution and employ backoff, where n-grams shorter than 10 are used to estimate the probability of infrequent 10-grams~\cite{chen1999empirical}.

The decoding operation finds the most likely sequence of hidden states in a Hidden Markov Model (HMM), where the emission probabilities are determined by the HWR model and the transition probabilities are determined by the LM:
\begin{equation} \label{eq:lm_decoding}
\mathbf{\hat{h}} = \arg\max_{\mathbf{h}}\prod_i^N p(h_i| h_{j < i}) p(x_i | h_i)^w
\end{equation}
where $\mathbf{h}$ is the sequence of hidden states corresponding to characters, $h_{j<i}$ indicates all states prior to $h_i$, $\boldsymbol{x}$ 
is the observed data, and $w$ determines the relative importance of the CNN-BLSTM and LM predictions.
Because characters can span multiple output frames, we model each character using 3 states (corresponding to character start, middle, and end) as is commonly done in speech recognition~\cite{mohri2008speech}.
The LM directly encodes the $p(h_i| h_{j < i})$ term, but the CNN-BLSTM outputs $p(h_i | x_i)$.
Using Bayes Rule, we have 
\begin{equation} \label{eq:bayes}
p(x_i | h_i) = \frac{p(h_i | x_i)p(x_i)}{p(h_i)}
\end{equation}

We can estimate $p(h_i)$ by examining the CNN-BLSTM outputs, but $p(x_i)$ is unknown.
Following Bluche et al., we approximate $\frac{p(x_i)}{p(h_i)} \approx p(h_i)^{-\alpha}$, where $\alpha$ is a hyperparameter~\cite{bluche2014comparison}.
An exact solution to Eq.~\ref{eq:lm_decoding} can be intractable, so in practice, we use a beam search which efficiently searches the state-space, but in some cases may not find the exact maximal sequence of characters.

\subsection{Hybrid Training}

\begin{algorithm}[t]
    \caption{Hybrid Training Procedure}
    \label{alg:hybrid}
    \begin{algorithmic}[1]
        \REQUIRE Source Model $\mathbf{S}$, Source Data $\mathbf{X_S}$, Source Transcriptions $\mathbf{Y_S}$, Target Data $\mathbf{X_T}$, Target LM $\mathbf{Q}$
        \ENSURE Target Model $\mathbf{T}$
        \STATE Initialize $\mathbf{T}$ with weights of $\mathbf{S}$
        \FOR{$k=0$ to $50$}
        		\STATE // First update LM marginal probabilities to reflect $\mathbf{T}$
        		\FOR{$j=0$ to $100$}    			
        			\STATE Sample a minibatch $\mathbf{x_T}$ from $\mathbf{X_T}$
        			\STATE $\mathbf{P} \leftarrow \mathbf{T}(\mathbf{x_T})$
        			\STATE Use $\mathbf{P}$ to update $p(h_i)$ (used in Eq.~\ref{eq:bayes})
        		\ENDFOR
        		\STATE // Train $\mathbf{T}$ on source and target data
        		\FOR{$j=0$ to $100$}    			
        			\STATE Sample a minibatch $\mathbf{x_T}$ from $\mathbf{X_T}$
         			\STATE $\mathbf{P} \leftarrow \mathbf{T}(\mathbf{x_T})$
        			\STATE $\mathbf{y_T} \leftarrow \mathbf{Q}(\mathbf{P})$ (Eq.~\ref{eq:lm_decoding})
        			
        			\STATE Sample a minibatch $(\mathbf{x_S},\mathbf{y_S})$ from $(\mathbf{X_S},\mathbf{Y_S})$
        			\STATE $\mathbf{x} \leftarrow \mathbf{x_T} \| \mathbf{x_S}$,~ $\mathbf{y} \leftarrow \mathbf{y_T} \| \mathbf{y_S}$ 
        			\STATE Update $\mathbf{T}$ according to $L(\mathbf{x}, \mathbf{y})$ (Eq.~\ref{eq:two_loss})
        		\ENDFOR
        \ENDFOR
    \end{algorithmic}
\end{algorithm}

Our hybrid training procedure leverages the recognition performance achieved by the source model on the source language to then learn recognition over the target language.
The overall process is shown in Algorithm~\ref{alg:hybrid}.

The main difference between hybrid and source training is in the data used for learning.
During hybrid training, part of the data comes from the source dataset (typically 50\%) with the rest coming from the target dataset for which there are no ground truth transcriptions.
However, the training loss for hybrid training is the same as in source training (Eq.~\ref{eq:two_loss}), which means that to train we need to provide some transcriptions for the target data.

We obtain target transcriptions by applying the LM of the target language to the predictions of the network.
The intuition is that due to the similarity of the source and target languages, the predictions of the network will be much better than random, though still quite poor at first.
Applying the LM will improve the poor predictions to make better targets, which in turn helps the network to learn the target language better.
We do, however, continue to train on source data to stabilize the learning process with real ground truth.

At the beginning of hybrid training, the model has never seen any instances of characters that are only part of the target language and will make incorrect predictions.
However, the LM can correct some of these errors based on the context of surrounding correct predictions.
For example, English words contain no accented characters, so a source model trained on English would never predict accented characters, but French and Spanish do use accents.
The LM is able to correct the model predictions to include accents and thus introduce these characters into the ground truth so the model can learn to predict these characters in the future.

Because LM decoding depends on the marginal distribution of CNN-BLSTM outputs, $p(h_i)$ in Eq.~\ref{eq:bayes}, we need to periodically update this quantity.
This is done in lines 3-7 in Algorithm~\ref{alg:hybrid}.
In normal HWR model training, this is unnecessary because the LM is applied only as post-processing and not as part of the training process.

\section{Datasets}


In this work we use 4 datasets: IAM~\cite{marti2002iam}, Rimes~\cite{Augustin2006}, Rodrigo~\cite{serrano2010rodrigo}, and Bentham (2014 HTR competition)~\cite{sanchez2014icfhr2014} collections.
Each dataset is composed of a number of line images with corresponding ground truth transcriptions.

Rodrigo is a single author, 853 page Spanish manuscript written in 1545 with 20000 segmented line images.
We used the first 750 pages as training data, the next 50 pages as validation data, and the remaining pages as test data.
The annotations contain some meta information that we preprocessed to exclude.
Some examples of this include symbols that indicate that whitespace should be inserted or deleted for correctness, i.e.~the manuscript author did not conform to modern usage of whitespace.

The Bentham collection are the writings of the English philosopher Jeremy Bentham (1748-1832), though some images may be handwritten copies of his works produced by others~\cite{sanchez2014icfhr2014}. 
For preprocessing, we deskewed the line images and performed height normalization.
For IAM, we use the standard split, merging the two defined validation sets.
For Rimes, there is only a defined train/test split, so we used a subset of the training data for a validation set.

Each image collection has different low level differences (e.g.~color, texture), so we opted to binarize each dataset to eliminate those differences.
This allows our analysis to focus on adapting to salient differences in language and style rather than on adapting to low level domain differences.
For IAM, Rimes, and Rodrigo, we used Otsu binarization~\cite{otsu1979threshold} but for Bentham, we used adaptive Wolf binarization~\cite{WolfICPR2002V} because it produced visually better binarizations.

To train the LMs for each dataset used in most experiments, we used the transcriptions of the training data.
Though this corresponds to having an optimal LM for hybrid training, we also explore using LMs trained on unrelated data.
For these LMs, we sampled 50000 sentences from the United Nations proceedings subset of the Europarl machine translation dataset~\cite{koehn2005europarl} in Spanish, English, and French.

To obtain the character classes predicted by our models, we take the union of the character sets of each dataset.
Because of this, during source model training, the classifiers output distributions over all characters, not just those characters contained in the source dataset.
This way if the target dataset has additional characters, we do not need to modify the classifiers before hybrid training. 

\section{Experimental Results}

\begin{table}[t]
\caption{CER of source models evaluated on each dataset.}
\label{tab:source}
\centering
\begin{tabular}{|ll|cccc|}
\hline
&& \multicolumn{4}{c|}{Test Data} \\
&& Bentham & IAM & Rimes & Rodrigo \\
\hline
\parbox[t]{1mm}{\multirow{3}{*}{\rotatebox[origin=c]{90}{Train Data}}}
& Bentham &  4.7 & 45.5 & 43.3 & 26.1 \\
& IAM     & 27.8 &  8.4 & 16.1 & 24.3 \\
& Rimes   & 35.0 & 24.6 &  4.9 & 34.4 \\
& Rodrigo & 69.5 & 67.0 & 67.1 &  6.8 \\
\hline

\end{tabular} 

\end{table}

\begin{table*}[t]
\caption{CER of hybrid trained models for all language pairs across a variety of experimental settings.}
\label{tab:hybrid}
\centering
\resizebox{2.08\columnwidth}{!}{
\begin{tabular}{|cccc|ccc|ccc|ccc|ccc|c|}
\hline
\multicolumn{4}{|c|}{Experimental Conditions} & \multicolumn{3}{|c|}{Source: Bentham} & \multicolumn{3}{|c|}{Source: IAM} & \multicolumn{3}{|c|}{Source: Rimes} & \multicolumn{3}{|c|}{Source Rodrigo} & \\
\hline
Source   & LM      & LM              & Amount    & \multicolumn{3}{|c|}{Transfer to} & \multicolumn{3}{|c|}{Transfer to} & \multicolumn{3}{|c|}{Transfer to} & \multicolumn{3}{|c|}{Transfer to} & Avg.\\
Epochs   & data    & $w$ / $\alpha$  & Source    &  IAM & Rim. & Rod. &  Ben. & Rim. & Rod. & Ben. &  IAM & Rod. & Ben. &  IAM & Rim. &  (no Rod.)\\
\hline
10       & Train   & 0.4 / 0.5       & 50\%      &  \textbf{9.2} &  \textbf{7.3} & \textbf{12.0} &  8.2  &  6.3 & 13.0 &  8.8 &  8.2 & \textbf{12.6} & 64.5 & 77.2 & 35.1 & 8.0  \\
50       & Train   & 0.4 / 0.5       & 50\%      &  9.8 &  7.5 & 13.3 &  8.6  &  6.3 & 13.5 &  9.3 &  8.4 & 13.4 & 70.3 & 74.3 & 70.1 & 8.3  \\
\hline
10       & Train   & 0.4 / 0.5       & 75\%      &  9.3 &  7.6 & 13.2 &  \textbf{7.5}  &  \textbf{5.7} & 12.1 &  9.4 &  \textbf{7.8} & 12.8 & \textbf{59.4} & \textbf{64.7} & 33.4 & \textbf{7.8}  \\
10       & Train   & 0.4 / 0.5       & 25\%      & 10.7 &  7.8 & 13.1 &  9.3  &  6.3 & \textbf{11.9} &  \textbf{8.5} &  8.7 & \textbf{12.6} & 71.8 & 73.8 & 37.5 & 8.6  \\
\hline
10       & Train   & 0.8 / 0.4       & 50\%      & 12.1 &  8.1 & 100.0&  8.2  &  6.7 & 97.0 &  8.9 &  8.9 & 13.5 & 67.2 & 80.1 & 18.3 & 8.8 \\
10       & Train   & 1.2 / 0.3       & 50\%      & 30.8 &  9.3 & 91.7 & 11.1  &  7.2 & 79.1 & 10.3 & 12.4 & 96.0 & 66.7 & 74.1 & \textbf{11.1} & 13.5\\
\hline
10       & None    & -               & 50\%      & 54.3 & 69.4 & 38.4 & 27.3  & 17.4 & 27.0 & 58.7 & 22.1 & 44.4 & 100.0& 84.4 & 98.4 & 41.5\\ 
10       & Europarl& 0.4 / 0.5       & 50\%      & 32.6 & 80.5 & 85.8 & 13.4  & 12.0 & 23.6 & 20.7 & 18.6 & 36.8 & 99.8 & 99.1 & 99.8 & 29.6\\
\hline
\multicolumn{4}{|c|}{Source Model - no Hybrid Training} & 45.5 & 43.3 & 26.1 & 27.8 & 16.1 & 24.3 & 35.0 & 24.6 & 34.4 & 59.5 & 67.0 & 67.1 & 32.1\\
\hline

\end{tabular} 
}
\end{table*}

In the following experiments, we use the following protocol.
For source models, we trained 4 models for each dataset for 10 epochs using the ADAM optimizer to perform weight updates~\cite{kingma2014adam}.
We then selected the best model using the Character Error Rate (CER) on the validation set after performing LM decoding using the dataset-specific LM.
All reported numbers for source models are on the designated test splits for each dataset.
These source models were used as the initial models in all hybrid training experiments, except where noted.

For hybrid training, we also trained 4 models where each hybrid model is initialized with the weights learned on the source dataset.
Hybrid models are trained for approximately 12000 weight updates using mini-batches of 8 images, where mini-batches contain both source and target images.
To report metrics, we select the best model based on the validation set for the target data and then evaluate this model on the target data.
While in practice this is not feasible because target data would not have ground truth transcriptions, this allowed us to fairly compare different methods of hybrid training.
We leave a method for selecting the best model without using ground truth as future work.

\subsection{Source Model Evaluation}

Table~\ref{tab:source} shows the CER of source models when evaluated on each dataset.
As expected, source models obtain low CER when the test data matches the training data and high CER when there is a mismatch.
Though this result may be obvious, it demonstrates the need for our hybrid training methodology in order to transfer models from one language to another.
We also note that even though IAM and Bentham are both English datasets, models trained on one do not perform well on the other and have need of transfer learning.

The CERs obtained are competitive when compared with previous results reported in the literature.
For example,~\cite{bluche2014comparison} reports CER of 3.9 and 3.8 for IAM and Rimes respectively, while we achieve 8.4 and 4.9 CERs.
In~\cite{granell2018transcription}, a CER of 3.0 is reported on the Rodrigo dataset, though this number is not directly comparable to our reported results because they use a different data split and transcription preprocessing.
Additionally, we binarized our data for transferability and generally CNN-BLSTM models perform better when using grayscale inputs.
The best CER on Bentham reported in the 2014 ICFHR HTR compeition is 5.0 for the restricted track~\cite{sanchez2014icfhr2014}.
Also, our reported numbers are on source models that have not trained to convergence (this improves hybrid training) but further training of the source models produces 1-2\% lower CERs.

\subsection{Hybrid Training}
\begin{figure*}
\centering
\begin{subfigure}[b]{0.3\textwidth} \centering
\includegraphics[width=\textwidth]{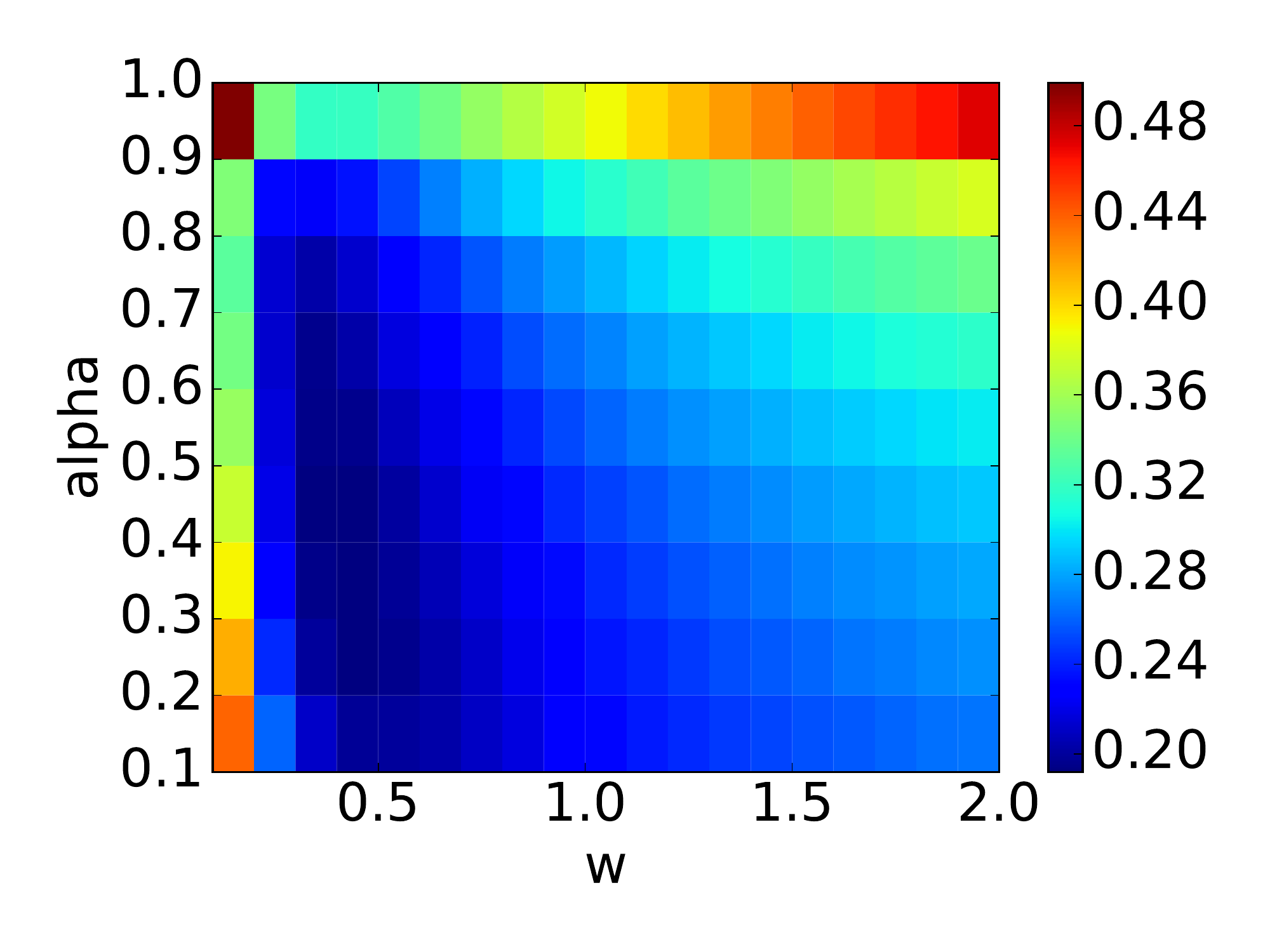}
\caption{Bentham}
\label{fig:lm_bentham}
\end{subfigure}\begin{subfigure}[b]{0.3\textwidth} \centering
\includegraphics[width=\textwidth]{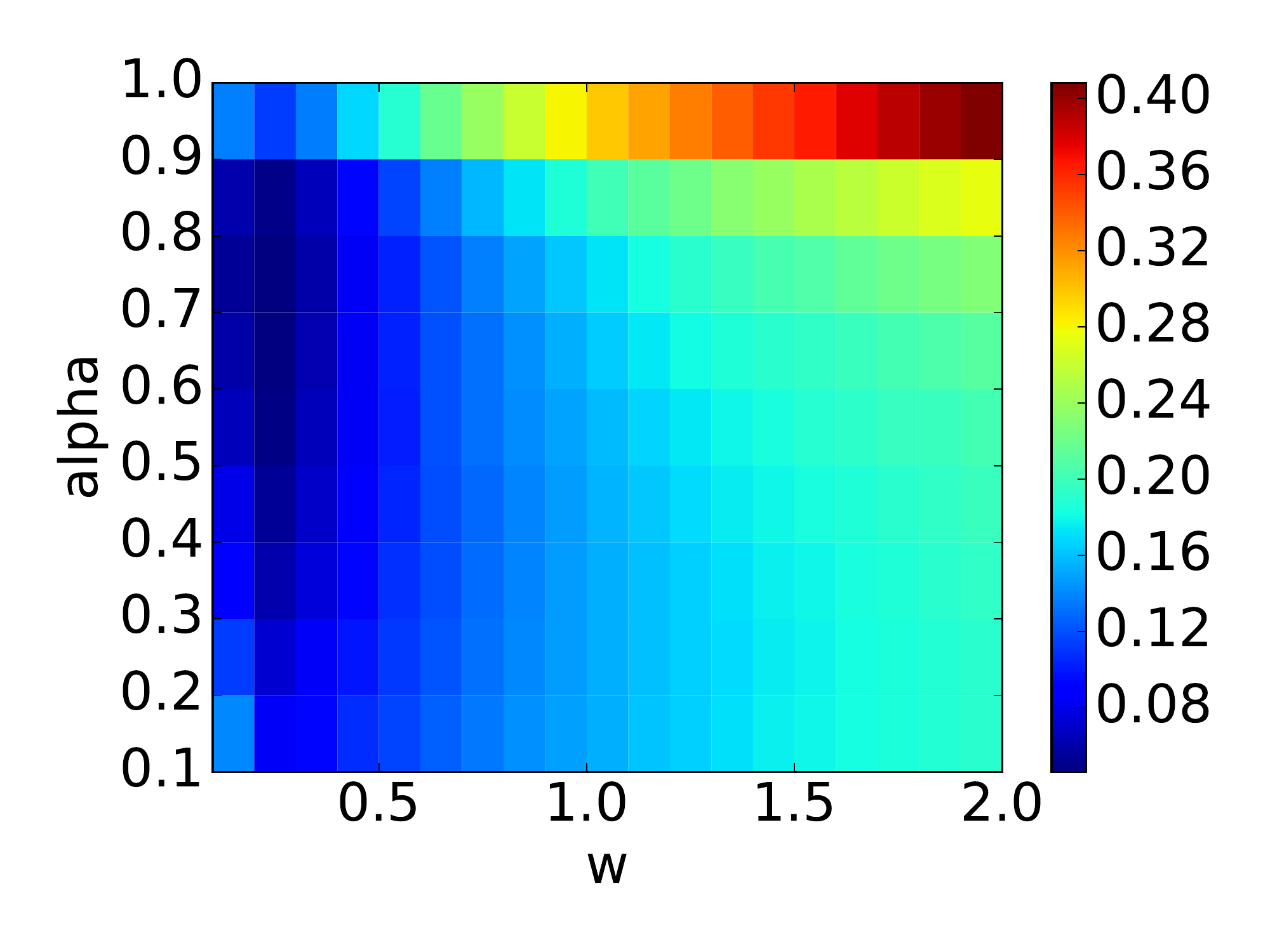}
\caption{Rimes}
\label{fig:lm_rimes}
\end{subfigure}\begin{subfigure}[b]{0.3\textwidth} \centering
\includegraphics[width=\textwidth]{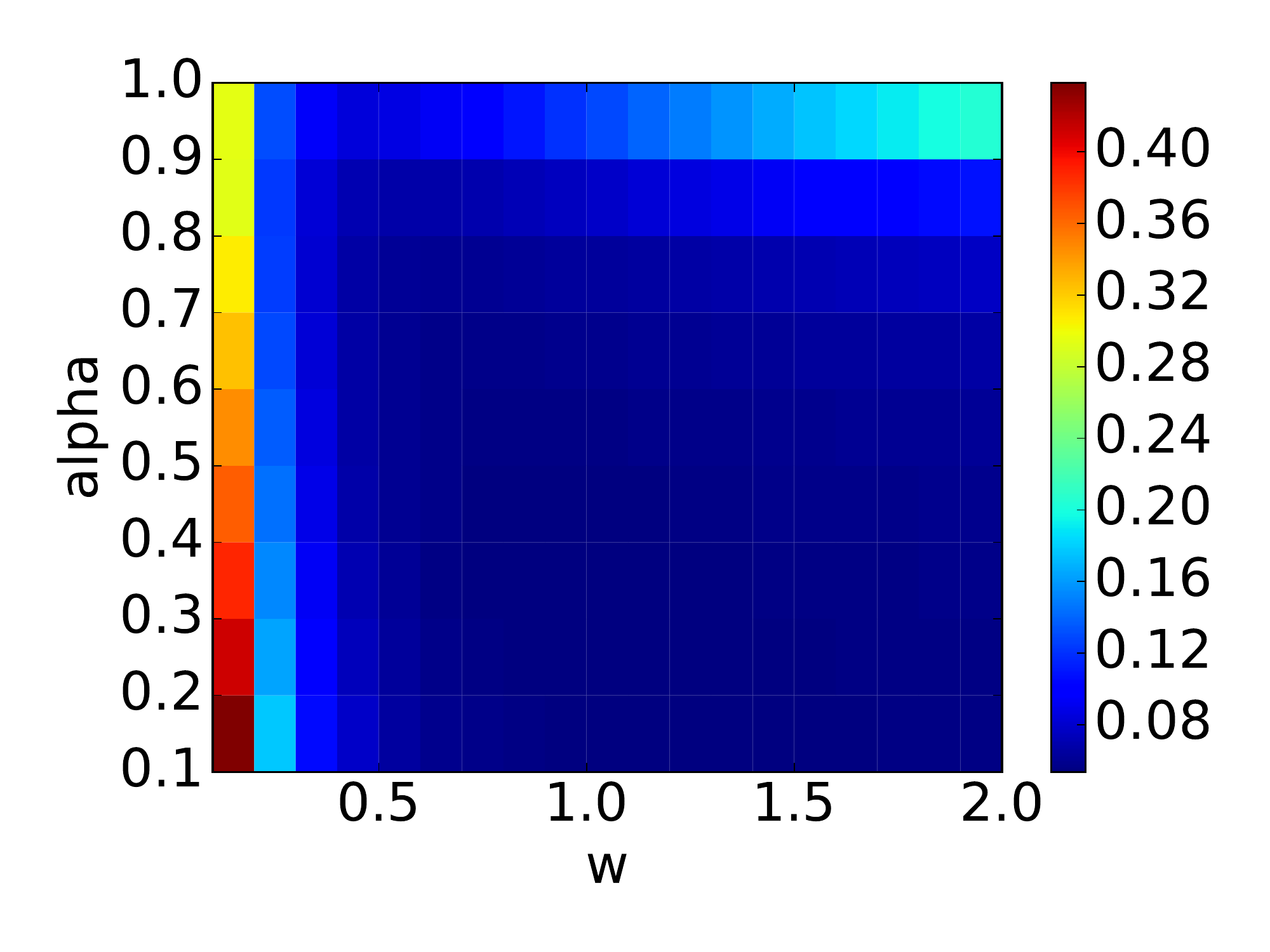}
\caption{IAM}
\label{fig:lm_iam}
\end{subfigure}

\caption{LM parameter search visualized as a heatmap of resulting CER.  We used the IAM source model and evaluated on Bentham, Rimes, and IAM validation sets for many values of $\alpha,w$. For Bentham and Rimes, the optimal $\alpha \approx 0.5$, $w \approx 0.4$.  For IAM, the optimal $\alpha=0.3$, $w=1.2$.}
\label{fig:lm_params}
\end{figure*}

In hybrid training, we varied 4 factors to gain a better understanding of the sensitivities of the method:
\begin{itemize}
\item Length of source model training time
\item Proportion of source and target data
\item Data used to train the LM
\item The $w$ and $\alpha$ LM parameters
\end{itemize}

Table~\ref{tab:hybrid} shows the CER after hybrid training for all language pairs for all experimental settings.
Here we explain the column semantics of Table~\ref{tab:hybrid}.
Source Epochs indicates how long source models were trained before hybrid training began.  
We also varied the data used for LM training - either the ground truth training set transcriptions, Europarl corpus subset, or no LM was used.  
The next 2 columns respectively indicate the LM hyperparameters and percentage of source data used in hybrid training.  
Remaining columns indicate Source-Target dataset pairs, where the first header row indicates the source language with target languages listed below.  
For example, the first data column is Bentham as the source with IAM as the target.  
The last column shows the average performance of the 6 language pairs involving Bentham, IAM, and Rimes.
For this average, we excluded Rodrigo because of the extremely high CERs of unsuccessful transfers, which would dominate the average.
For comparison, the last row shows performance of the source models before hybrid training, i.e.~the off-diagonal entries of Table~\ref{tab:source}.

Considering the first 4 rows of Table~\ref{tab:hybrid}, pairwise transfers between Bentham, IAM, and Rimes are extremely successful, achieving CERs near to those obtained with full supervised training in some cases.
It is interesting that while these three datasets can transfer to Rodrigo with CERs of about 13\%, the reverse is not true.
Only Rodrigo to Rimes hybrid training managed to significantly improve the CER over the source model, achieving 11.1\% CER under one set of experimental conditions, though this language pair appears sensitive to variations in experimental conditions.

When the source model is trained to convergence, i.e.~trained for 50 epochs instead of 10, CER on the source data improves by about 1-2\% (data not shown), but the CER after transferring increases for all language pairs except one.
The average CER increases by 0.3\%.
This is because after training for so long, the models can overfit the source data and may have difficulty unlearning factors unique to the source dataset.

Next we examined what proportion of source and target data is used during hybrid training.
Overall, using 75\% source data produces an average CER of 7.8\% vs 8.0\% for equal proportions and 8.6\% for 25\% source data.
We also note that the optimal percentage of source data varies by language pair.
Because the target labels provided by the LM are not always correct, the model can diverge if it is presented with too many poor quality target labels.
Source data helps stabilize hybrid training, so using a larger proportion of source data may make training more stable.

Next we examined the LM parameters $w$ and $\alpha$ used during LM decoding (Eqs.~\ref{eq:lm_decoding},\ref{eq:bayes}).
We determined our default values of $w=0.4$ and $\alpha=0.5$ by cross validation to optimize the CER of source models evaluated on the datasets that they were not trained on (i.e., the off-diagonal entries of Table~\ref{tab:source}).
For example, Fig.~\ref{fig:lm_params} shows heatmaps for the source IAM model evaluated on Bentham and Rimes.
When evaluating the IAM model on Bentham or Rimes, we see better performance when $w \approx 0.4$ and $\alpha \approx 0.5$, but when we evaluate on IAM, $w=1.2$ and $\alpha=0.3$ perform best.
We saw a similar trend when evaluating the Bentham source model on the other datasets.

A similar trend also holds for hybrid training.
Our default parameters of $w=0.4$, $\alpha=0.5$ achieved an average CER of 8.0\%, which is lower than 8.6\% with $w=0.8$, $\alpha=0.4$ and 13.5 with $w=1.2$, $\alpha=0.3$.
Also, transferring to Rodrigo becomes unsuccessful when using these alternate parameters.
However, it is interesting that these parameter settings greatly improve transfer from Rodrigo to Rimes (achieving 18.3 and 11.1 CER).
Thus the optimal LM hyperparameters vary based on the language pair, and unfortunately, they cannot be estimated by cross validation in a real setting as cross validation relies on the ground truth for the target language.

We conclude our experiments by varying the data used to train the LM.
If we do not apply LM decoding during hybrid training (or equivalently use a LM where all sequences of characters are equally likely), we see that some language pairs  improve over the source model performance, though some do not improve or get worse.
When we use the Europarl trained LMs, we see degraded performance with respect to the LMs trained on the dataset training sets, but this is expected to some degree.
The Europarl corpus uses very formal language, and the modern Spanish is very different from the historical Spanish used in the 1545 Rodrigo manuscript.
Transferring Bentham to IAM and vice versa, using the Europarl English LM greatly improves CER compared to using no LM at all.
The same is also true for IAM and Rimes.

\section{Conclusion}

In this work we proposed a methodology that trains HWR on a target language without using any labeled data in that language.
It does so by leveraging labeled images in a closely related source language and a language model in the target language.
After training a source model, we train on both the source and target data, inputting target labels using the current model predictions decoded by the LM.
We demonstrate that our approach is successful on many pairs of languages using the IAM, Rimes, Bentham, and Rodrigo datasets.
We explored the design choices of our hybrid training approach and make conclusions about the LM training data, LM hyperparameters, amount of source data in hybrid training, and length of source model training.

{
\newcommand{\BIBdecl}{\setlength{\itemsep}{0.25 em}}
\bibliographystyle{IEEEtran}
\bibliography{bib}

\begin{thebibliography}{10}
\providecommand{\url}[1]{#1}
\csname url@samestyle\endcsname
\providecommand{\newblock}{\relax}
\providecommand{\bibinfo}[2]{#2}
\providecommand{\BIBentrySTDinterwordspacing}{\spaceskip=0pt\relax}
\providecommand{\BIBentryALTinterwordstretchfactor}{4}
\providecommand{\BIBentryALTinterwordspacing}{\spaceskip=\fontdimen2\font plus
\BIBentryALTinterwordstretchfactor\fontdimen3\font minus
  \fontdimen4\font\relax}
\providecommand{\BIBforeignlanguage}[2]{{%
\expandafter\ifx\csname l@#1\endcsname\relax
\typeout{** WARNING: IEEEtran.bst: No hyphenation pattern has been}%
\typeout{** loaded for the language `#1'. Using the pattern for}%
\typeout{** the default language instead.}%
\else
\language=\csname l@#1\endcsname
\fi
#2}}
\providecommand{\BIBdecl}{\relax}
\BIBdecl

\bibitem{graves2009novel}
A.~Graves, M.~Liwicki, S.~Fern{\'a}ndez, R.~Bertolami, H.~Bunke, and
  J.~Schmidhuber, ``A novel connectionist system for unconstrained handwriting
  recognition,'' \emph{IEEE transactions on pattern analysis and machine
  intelligence}, vol.~31, no.~5, pp. 855--868, 2009.

\bibitem{simard2003best}
P.~Y. Simard, D.~Steinkraus, J.~C. Platt \emph{et~al.}, ``Best practices for
  convolutional neural networks applied to visual document analysis.'' in
  \emph{ICDAR}, vol.~3, 2003, pp. 958--962.

\bibitem{varga2008perturbation}
T.~Varga and H.~Bunke, ``Perturbation models for generating synthetic training
  data in handwriting recognition,'' in \emph{Machine Learning in Document
  Analysis and Recognition}.\hskip 1em plus 0.5em minus 0.4em\relax Springer,
  2008, pp. 333--360.

\bibitem{wigington2017data}
C.~Wigington, S.~Stewart, B.~Davis, B.~Barrett, B.~Price, and S.~Cohen, ``Data
  augmentation for recognition of handwritten words and lines using a cnn-lstm
  network,'' in \emph{ICDAR}, vol.~1.\hskip 1em plus 0.5em minus 0.4em\relax
  IEEE, 2017, pp. 639--645.

\bibitem{tung1994performance}
C.-H. Tung, Y.-J. Chen, and H.-J. Lee, ``Performance analysis of an ocr system
  via an artificial handwritten chinese character generator,'' \emph{Pattern
  Recognition}, vol.~27, no.~2, pp. 221--232, 1994.

\bibitem{lee1998new}
D.-H. Lee and H.-G. Cho, ``A new synthesizing method for handwriting korean
  scripts,'' \emph{International Journal of Pattern Recognition and Artificial
  Intelligence}, vol.~12, no.~01, pp. 45--61, 1998.

\bibitem{elarian2015arabic}
Y.~Elarian, I.~Ahmad, S.~Awaida, W.~G. Al-Khatib, and A.~Zidouri, ``An arabic
  handwriting synthesis system,'' \emph{Pattern Recognition}, vol.~48, no.~3,
  pp. 849--861, 2015.

\bibitem{frinken2011co}
V.~Frinken, A.~Fischer, H.~Bunke, and A.~Foornes, ``Co-training for handwritten
  word recognition,'' in \emph{ICDAR}.\hskip 1em plus 0.5em minus 0.4em\relax
  IEEE, 2011, pp. 314--318.

\bibitem{ball2009semi}
G.~R. Ball and S.~N. Srihari, ``Semi-supervised learning for handwriting
  recognition,'' in \emph{ICDAR}.\hskip 1em plus 0.5em minus 0.4em\relax IEEE,
  2009, pp. 26--30.

\bibitem{graves2006connectionist}
A.~Graves, S.~Fern{\'a}ndez, F.~Gomez, and J.~Schmidhuber, ``Connectionist
  temporal classification: labelling unsegmented sequence data with recurrent
  neural networks,'' in \emph{International Conference on Machine
  learning}.\hskip 1em plus 0.5em minus 0.4em\relax ACM, 2006, pp. 369--376.

\bibitem{Povey_ASRU2011}
D.~Povey, A.~Ghoshal, G.~Boulianne, L.~Burget, O.~Glembek, N.~Goel,
  M.~Hannemann, P.~Motlicek, Y.~Qian, P.~Schwarz, J.~Silovsky, G.~Stemmer, and
  K.~Vesely, ``The kaldi speech recognition toolkit,'' in \emph{IEEE 2011
  Workshop on Automatic Speech Recognition and Understanding}, Dec. 2011.

\bibitem{bluche2014comparison}
T.~Bluche, H.~Ney, and C.~Kermorvant, ``A comparison of sequence-trained deep
  neural networks and recurrent neural networks optical modeling for
  handwriting recognition,'' in \emph{International Conference on Statistical
  Language and Speech Processing}.\hskip 1em plus 0.5em minus 0.4em\relax
  Springer, 2014, pp. 199--210.

\bibitem{voigtlaender2016handwriting}
P.~Voigtlaender, P.~Doetsch, and H.~Ney, ``Handwriting recognition with large
  multidimensional long short-term memory recurrent neural networks,'' in
  \emph{ICFHR}.\hskip 1em plus 0.5em minus 0.4em\relax IEEE, 2016, pp.
  228--233.

\bibitem{chen1999empirical}
S.~F. Chen and J.~Goodman, ``An empirical study of smoothing techniques for
  language modeling,'' \emph{Computer Speech \& Language}, vol.~13, no.~4, pp.
  359--394, 1999.

\bibitem{mohri2008speech}
M.~Mohri, F.~Pereira, and M.~Riley, ``Speech recognition with weighted
  finite-state transducers,'' in \emph{Springer Handbook of Speech
  Processing}.\hskip 1em plus 0.5em minus 0.4em\relax Springer, 2008, pp.
  559--584.

\bibitem{marti2002iam}
U.-V. Marti and H.~Bunke, ``The iam-database: an english sentence database for
  offline handwriting recognition,'' \emph{IJDAR}, vol.~5, no.~1, pp. 39--46,
  2002.

\bibitem{Augustin2006}
E.~Augustin, J.-m. Brodin, M.~Carré, E.~Geoffrois, E.~Grosicki, and
  F.~Prêteux, ``{RIMES evaluation campaign for handwritten mail processing},''
  in \emph{IWFHR}, 2006.

\bibitem{serrano2010rodrigo}
N.~Serrano, F.~Castro, and A.~Juan, ``The rodrigo database.'' in \emph{LREC},
  2010, pp. 19--21.

\bibitem{sanchez2014icfhr2014}
J.~A. S{\'a}nchez, V.~Romero, A.~H. Toselli, and E.~Vidal, ``Icfhr2014
  competition on handwritten text recognition on transcriptorium datasets
  (htrts),'' in \emph{ICFHR}.\hskip 1em plus 0.5em minus 0.4em\relax IEEE,
  2014, pp. 785--790.

\bibitem{otsu1979threshold}
N.~Otsu, ``A threshold selection method from gray-level histograms,''
  \emph{IEEE transactions on systems, man, and cybernetics}, vol.~9, no.~1, pp.
  62--66, 1979.

\bibitem{WolfICPR2002V}
C.~Wolf, J.-M. Jolion, and F.~Chassaing, ``Text {L}ocalization, {E}nhancement
  and {B}inarization in {M}ultimedia {D}ocuments,'' in \emph{Proceedings of the
  {I}nternational {C}onference on {P}attern {R}ecognition}, vol.~2, 2002, pp.
  1037--1040.

\bibitem{koehn2005europarl}
P.~Koehn, ``Europarl: A parallel corpus for statistical machine translation,''
  in \emph{MT summit}, vol.~5, 2005, pp. 79--86.

\bibitem{kingma2014adam}
D.~P. Kingma and J.~Ba, ``Adam: A method for stochastic optimization,''
  \emph{arXiv preprint arXiv:1412.6980}, 2014.

\bibitem{granell2018transcription}
E.~Granell, E.~Chammas, L.~Likforman-Sulem, C.-D. Mart{\'\i}nez-Hinarejos,
  C.~Mokbel, and B.-I. C{\^\i}rstea, ``Transcription of spanish historical
  handwritten documents with deep neural networks,'' \emph{Journal of Imaging},
  vol.~4, no.~1, p.~15, 2018.

\end{thebibliography}

}
	
\end{document}